\documentclass[10pt,twocolumn,letterpaper]{article}

\usepackage{cvpr}              

\usepackage{amstext} 
\usepackage{array}   
\newcolumntype{L}{>{$}l<{$}} 
\usepackage{cuted}
\usepackage{amsmath}
\usepackage{amsfonts}
\usepackage{amssymb}

\definecolor{cvprblue}{rgb}{0.21,0.49,0.74}
\usepackage[pagebackref,breaklinks,colorlinks,allcolors=cvprblue]{hyperref}

\def\paperID{*****} 
\def\confName{CVPR}
\def\confYear{2025}

\title{AIM 2025 Rip Current Segmentation (RipSeg) Challenge Report}

\author{
Andrei Dumitriu$^{1, 2}$ \and Florin Miron$^{2}$ \and Florin Tatui$^{2}$ \and  Radu Tudor Ionescu$^2$ \and Radu Timofte$^1$ \and Aakash Ralhan$^{1}$ \and Florin-Alexandru Vasluianu$^1$ \and Shenyang Qian \and Mitchell Harley \and Imran Razzak \and Yang Song \and Pu Luo \and Yumei Li \and Cong Xu \and Jinming Chai \and Kexin Zhang \and Licheng Jiao \and Lingling Li \and Siqi Yu \and Chao Zhang \and Kehuan Song \and Fang Liu \and Puhua Chen \and Xu Liu \and Jin Hu \and Jinyang Xu \and Biao Liu\\ $^{1}$Computer Vision Lab, CAIDAS \& IFI, University of Würzburg, Germany \\
$^{2}$University of Bucharest, Romania\\
{\tt\small {andrei.dumitriu}@uni-wuerzburg.de}\\
}

\begin{document}

\twocolumn[{%
\renewcommand\twocolumn[1][]{#1}%
\maketitle
\begin{center}
    \centering
    \captionsetup{type=figure}
    
    \parbox{0.246\textwidth}{\centering Aerial - Bird's Eye}%
    \hfill
    \parbox{0.246\textwidth}{\centering Aerial - Tilted}%
    \hfill
    \parbox{0.246\textwidth}{\centering Elevated Beachfront}%
    \hfill
    \parbox{0.246\textwidth}{\centering Water-Level Beachfront}%
    
    \includegraphics[width=0.246\textwidth]{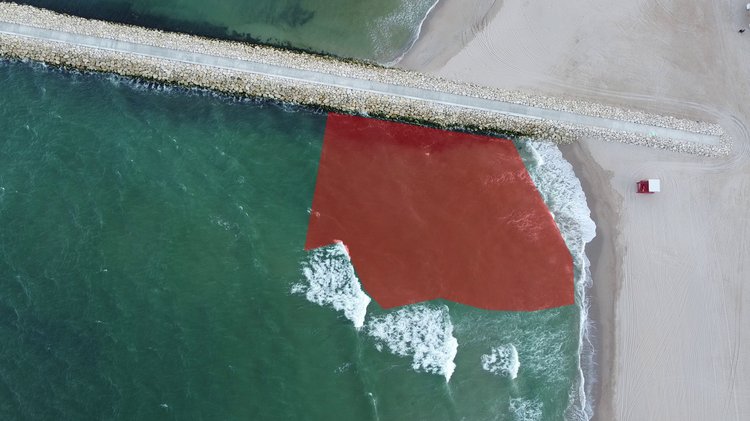}
    \includegraphics[width=0.246\textwidth]{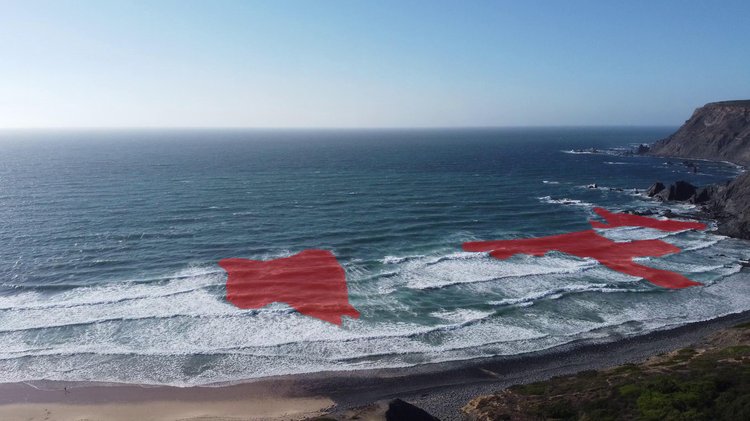}
    \includegraphics[width=0.246\textwidth]{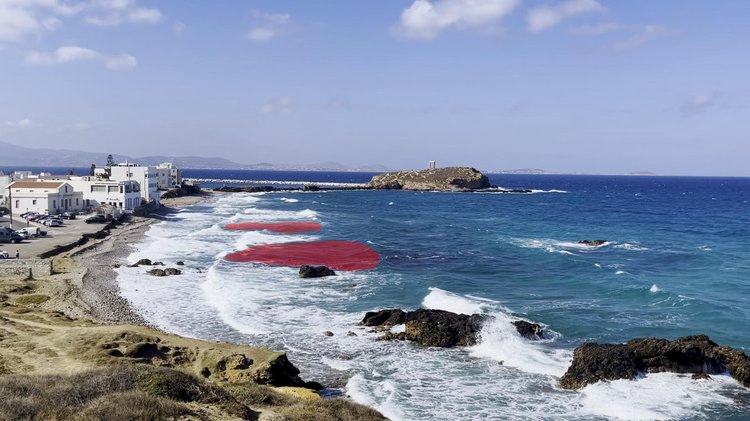}
    \includegraphics[width=0.246\textwidth]{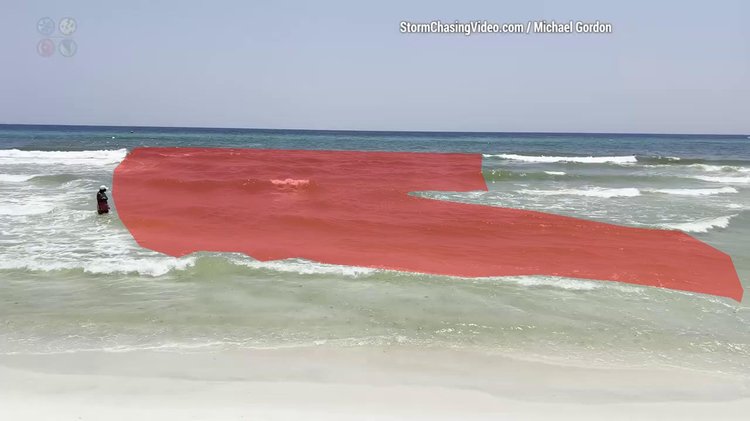}
\end{center}%
\vspace{-7mm} 
\begin{center}
    \centering
    \captionsetup{type=figure}
    \includegraphics[width=0.246\textwidth]{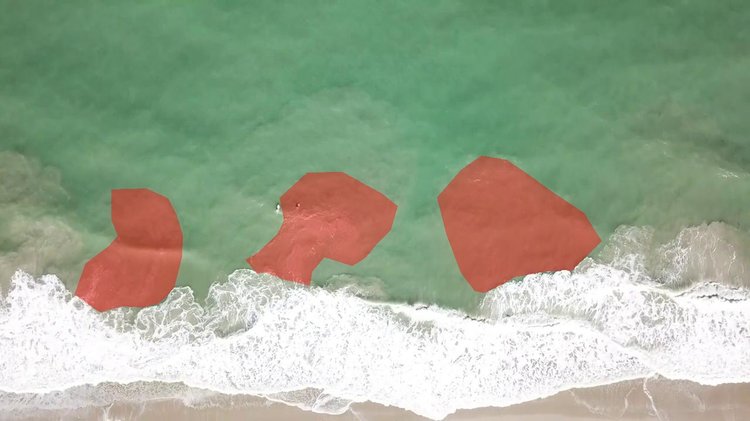}
    \includegraphics[width=0.246\textwidth]{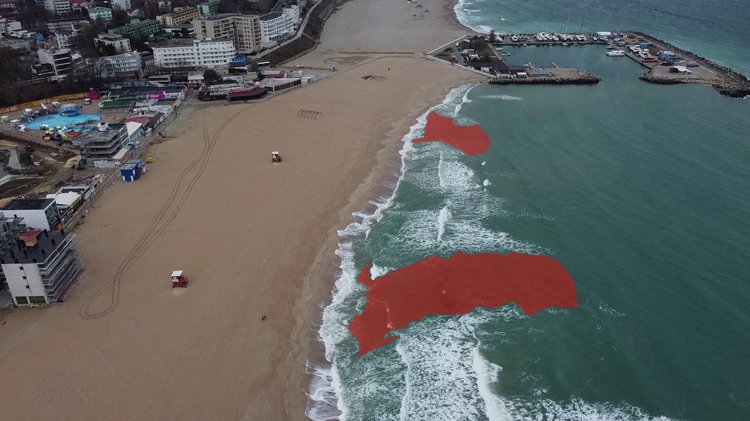}
    \includegraphics[width=0.246\textwidth]{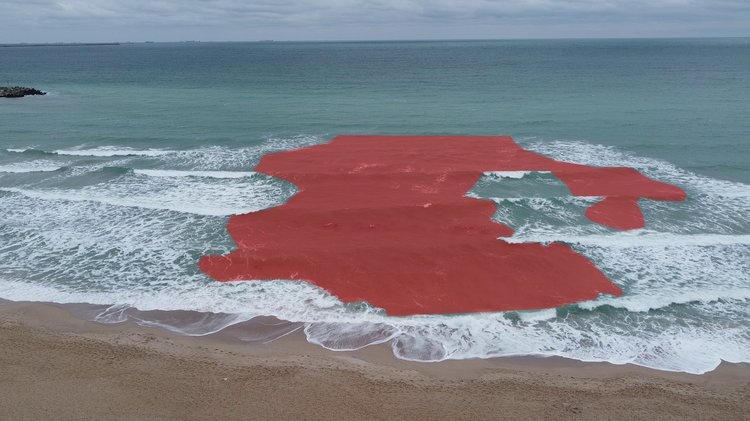}
    \includegraphics[width=0.246\textwidth]{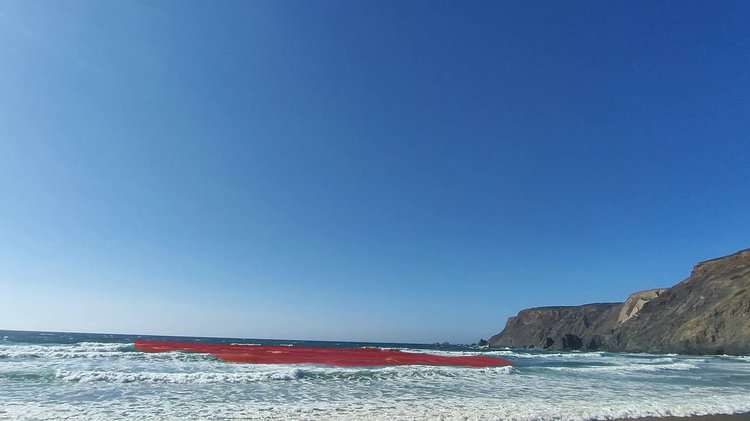}
    \vspace{-0.4cm}
    \captionof{figure}{Examples from the RipVIS dataset \cite{Dumitriu_2025_CVPR}, which also forms the basis of the RipSeg Challenge. The four columns illustrate different camera orientations: (a) aerial bird’s-eye, (b) aerial tilted, (c) elevated beachfront, and (d) water-level beachfront. The examples highlight the diversity of rip currents across locations, types, and viewpoints. Rip currents are visible through disrupted wave-breaking patterns, sediment transport, and deflection flows, with annotations shown in red. Best viewed in color.}
\label{fig:intro_presentation}
\end{center}%
}]
\maketitle

\begin{abstract}

This report presents an overview of the AIM 2025 RipSeg Challenge, a competition designed to advance techniques for automatic rip current segmentation in still images. Rip currents are dangerous, fast-moving flows that pose a major risk to beach safety worldwide, making accurate visual detection an important and underexplored research task. The challenge builds on RipVIS, the largest available rip current dataset, and focuses on single-class instance segmentation, where precise delineation is critical to fully capture the extent of rip currents. The dataset spans diverse locations, rip current types, and camera orientations, providing a realistic and challenging benchmark.

In total, $75$ participants registered for this first edition, resulting in $5$ valid test submissions. Teams were evaluated on a composite score combining $F_1$, $F_2$, $AP_{50}$, and $AP_{[50:95]}$, ensuring robust and application-relevant rankings. The top-performing methods leveraged deep learning architectures, domain adaptation techniques, pretrained models, and domain generalization strategies to improve performance under diverse conditions.

This report outlines the dataset details, competition framework, evaluation metrics, and final results, providing insights into the current state of rip current segmentation. We conclude with a discussion of key challenges, lessons learned from the submissions, and future directions for expanding RipSeg.

\end{abstract}
{\let\thefootnote\relax\footnotetext{%
\hspace{-5mm} 
$^{1, 2}$
Andrei Dumitriu, Aakash Ralhan, 
Florin Miron, Florin Tatui, Florin-Alexandru Vasluianu,
Radu Tudor Ionescu and Radu Timofte are the ICCV 2025 
challenge organizers. The other authors participated in the challenge.\\
\url{https://cvlai.net/aim/2025/}
}}

\section{Introduction}
\label{sec:intro}

\begin{table*}[t!]
    \centering
    \begin{tabular}{|c|ll|cccc|c|}
        \hline
        Rank & Team Name & Username &
        $F_1\uparrow$ & 
        $F_2\uparrow$ & 
        $AP_{50}\uparrow$ & 
        $AP_{[50:95]}\uparrow$ & 
        Final Score$\uparrow$ \\
        \hline
        1 & RipEye & shenyang115 & 0.72 & 0.74 & 0.69 & 0.33 & 0.68 \\
        2 & RipSense & aakash & 0.71 & 0.70 & 0.66 & 0.27 & 0.65 \\
        3 & Gogogochufalou & luopuu & 0.68 & 0.72 & 0.64 & 0.25 & 0.64 \\
        4 & ZYS & zhangcc & 0.67 & 0.65 & 0.45 & 0.17 & 0.55 \\
        5 & Simplehh & gl0ria & 0.65 & 0.66 & 0.43 & 0.16 & 0.54 \\
        \hline
    \end{tabular}
    \caption{Quantitative results of the solutions in the final phase. The value for the final score is provided, alongside the values for $F_1$, $F_2$, $AP_{50}$ and $AP_{[50:95]}$. The final score is computed as $\texttt{score} = 0.3 \cdot F_1 + 0.3 \cdot F_2 + 0.3 \cdot AP_{50} + 0.1 \cdot AP_{[50:95]}.$ Results are computed on the AIM2025 Rip Current Segmentation (RipSeg) test split.} 
    \label{tab:test_results} 
\end{table*}

Rip currents are powerful, fast-moving surface flows that pull water seaward from the shore, posing a significant hazard to swimmers and other beachgoers worldwide \cite{da2003analysis, lushine1991study, brewster2019estimations, brander2013brief}. Found along ocean, sea, and even large lake coastlines, their strength and shape are influenced by local hydrodynamics, seabed morphology, and in some cases, human-made coastal structures \cite{brander2000morphodynamics, castelle2016rip}. At peak speeds exceeding $8.7$ km/h (faster than an Olympic swimmer), they can swiftly carry individuals offshore \cite{noaa2023ripcurrents}. The danger is amplified by the fact that many swimmers fail to recognize the phenomenon and instinctively attempt to fight against the current, leading to exhaustion and potentially fatal outcomes. Public safety campaigns recommend swimming parallel to shore to escape, but this advice only helps if the current is correctly identified beforehand, ideally through proactive detection and warning systems.

In recent years, advances in computer vision have significantly improved capabilities in object detection, segmentation, and classification \cite{he2017mask, Jocher_YOLO_by_Ultralytics_2023, kirillov2023segment, ravi2024sam}, enabled in large part by high-quality datasets such as COCO \cite{lin2014microsoft}, Cityscapes \cite{cordts2016cityscapes}, and YouTube-VIS \cite{Yang2019vis, vis2021}. Initially, automatic rip current identification has been fragmented and slow compared with the advances in computer vision, but this field has recently gained more traction among computer vision scientists as well \cite{choi2024explainable, dumitriu2023rip, de2023ripviz, rashid2023reducing, zhu2022yolo, mori2022flow, mcgill2022flow, rampal2022interpretable, desilva2021frcnn, rashid2021ripdet, rashid2020ripnet, maryan2019machine, philip2016flow, Dumitriu_2025_CVPR, khan2025ripscout, khan2025ripfinder}. Due to the amorphous nature of rip currents, their high variety of types, the distinct camera orientations and the diversity in the natural environment where they can occur, automatic rip current identification remains a highly challenging task. Rip currents lack rigid boundaries, often manifesting through subtle visual cues such as disrupted wave patterns, sediment plumes, or localized changes in water color. These cues are dynamic, environment-dependent, and easily obscured by camera perspective, lighting, and weather conditions.

The introduction of RipVIS \cite{Dumitriu_2025_CVPR}, the largest publicly available rip current dataset, marked a significant step forward by providing a large-scale video instance segmentation benchmark dedicated to rip current detection. RipVIS originally focused on a single task, namely instance segmentation, which is perhaps the most valuable of the possible vision task formulations for rip current analysis. Unlike bounding box detection, which can include large amounts of irrelevant background, or classification, which offers no localization, instance segmentation can accurately delineate the full extent of a rip current without omitting significant parts or including unrelated regions.

Building on RipVIS, we introduce the RipSeg Challenge at ICCV 2025, the first competition dedicated to rip current segmentation in still images. As this is the inaugural edition, we focus exclusively on still images rather than videos. To that end, we use all available annotated frames from RipVIS \cite{Dumitriu_2025_CVPR}, which was originally annotated at a varied sampling rate, along with the training data from Dumitriu~\etal \cite{dumitriu2023rip}, resulting in a total of 27,718 images, split among 4 camera orientations, presented in Figure \ref{fig:intro_presentation}. These are divided into 18,386 for training, 4,348 for validation, and 4,984 for testing. The challenge targets single-class (rip current) instance segmentation, where multiple instances may occur in a single image, encouraging models that excel at fine-grained, pixel-level delineation in complex aquatic environments. By providing a carefully curated split and an unseen test set, RipSeg promotes methods that generalize well, while mitigating overfitting risks. We also define a custom weighted evaluation metric that balances standard segmentation measures with application-driven priorities, reflecting the real-world need for both accuracy and reliability in beach safety monitoring systems.

This challenge is one of the AIM 2025\footnote{\url{https://www.cvlai.net/aim/2025/}} workshop associated challenges on: high FPS non-uniform motion deblurring~\cite{aim2025highfps}, rip current segmentation~\cite{aim2025ripseg}, inverse tone mapping~\cite{aim2025tone}, robust offline video super-resolution~\cite{aim2025videoSR}, low-light raw video denoising~\cite{aim2025videodenoising}, screen-content video quality assessment~\cite{aim2025scvqa}, real-world raw denoising~\cite{aim2025rawdenoising}, perceptual image super-resolution~\cite{aim2025perceptual}, efficient real-world deblurring~\cite{aim2025efficientdeblurring}, 4K super-resolution on mobile NPUs~\cite{aim20254ksr}, efficient denoising on smartphone GPUs~\cite{aim2025efficientdenoising}, efficient learned ISP on mobile GPUs~\cite{aim2025efficientISP}, and stable diffusion for on-device inference~\cite{aim2025sd}.

\begin{figure*}[t]
    \centering
    \includegraphics[width=0.75\textwidth]{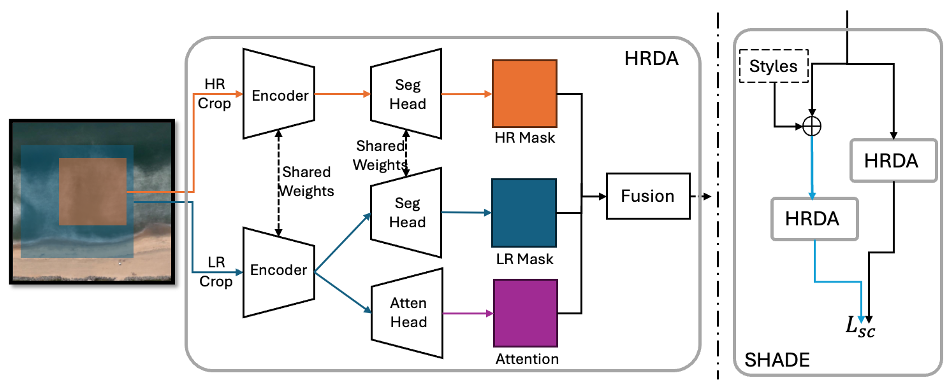}
    \caption{Methodology of RipEye, using HRDA \cite{hoyer2022daformer} and SHADE \cite{zhao2022style}.}
    \label{fig:RipEye}
\end{figure*}

\section{Challenge: Format and Ranking}
\label{sec:eval}
The AIM 2025 Rip Current Segmentation (RipSeg) Challenge was organized on CodaBench \cite{codabench} and was split into 3 phases: in the first phase (training), users were provided with a set of $18,386$ images for training, along with the polygon annotations in COCO JSON format. Participants had about two weeks to familiarize themselves with the task, format and evaluation criteria. The training phase was followed by a validation phase, where participants were provided with an extra set of $4,348$ images, along with a COCO JSON without annotations, to ensure correct prediction format. In this phase, participants could submit their predictions on the Codabench server and see their results, when compared to the private ground-truth annotations. Participants had around $5$ weeks for this phase, where they could develop and fine tune their models. In the last phase (test), the participants were provided with $4,984$ images with no annotations, and they had 2 days to submit their predictions (only one try), in order to see their final score. The short time provided for the test phase was intentional, in order to minimize risk of fraud by manually annotating the images. In validation phases, structure could be derived from the filenames, i.e.~files containing rip currents used the naming scheme of RipSeg-\texttt{<number>}, while files without rip currents were named RipSeg-NR-\texttt{<number>}. For the final test set, all files have been renamed with a randomized hash as an extra step in preventing fraud. After the two days, participants had to submit a description of their team, their reproducible code and a description of their approach. Only teams that passed this final check were considered in the final competition ranking, presented in Table \ref{tab:test_results}. The evaluation code was made publicly available since the beginning, for transparency.

Correct rip current identification is a safety critical task, where the $F_2$ score is one of the relevant metrics. While $F_2$ is useful in real-world scenarios, in a challenge format, the score can be increased by preferring an increased number of false positives. In order to mitigate this, we employed a weighted average of four relevant metrics in image segmentation, namely $F_1$, $F_2$, $AP_{50}$ and $AP_{[50:95]}$. More precisely, the final ranking was established considering the weighted average computed via the following formula:
\begin{equation}
\label{eq:final_score}
\texttt{score} = 0.3 \cdot F_1 + 0.3 \cdot F_2 + 0.3 \cdot AP_{50} + 0.1 \cdot AP_{[50:95]}
\end{equation}


\section{Methods}
\label{sec:methods}

The affiliations of challenge organizers and participants are included in Table \ref{tab:teams_affil}. We next present the approach submitted by each team.

\begin{table*}[t!]
\centering
\renewcommand{\arraystretch}{1.1}
\setlength{\tabcolsep}{3pt}
\begin{tabular}{|p{3.3cm}|p{4.7cm}|p{5.5cm}|p{3cm}|}
\hline
{Team} & {Members} & {Affiliations} & Contact\\
\hline

AIM 2025 Rip Current Segmentation (RipSeg) & 
Andrei Dumitriu$^{1,2}$, Florin Miron$^{2}$, Florin Tatui$^{2}$, Radu Tudor Ionescu$^{2}$, Radu Timofte$^{1}$, Aakash Ralhan$^{1}$, Florin-Alexandru Vasluianu$^{1}$&
$^1$ Computer Vision Lab, IFI \& CAIDAS, University of W\"urzburg, Germany \newline
$^2$ University of Bucharest, Romania & andrei.dumitriu@uni-wuerzburg.de \\
\hline
RipEye & 
Shenyang Qian$^3$, Mitchell Harley$^4$, Imran Razzak$^5$, Yang Song$^3$ &
$^3$ School of Computer Science and Engineering, UNSW Sydney, Australia \newline
$^4$ School of Civil and Environmental Engineering, UNSW Sydney, Australia \newline
$^5$ Mohamed bin Zayed University of Artificial Intelligence, Abu Dhabi, UAE & shenyang.qian@stud ent.unsw.edu.au \\
\hline

RipSense & 
Aakash Ralhan$^1$ & 
$^1$ Computer Vision Lab, IFI \& CAIDAS, University of W\"urzburg, Germany & aakash.ralhan@stud-mail.uni-wuerzburg.de \\
\hline

Gogogochufalou & 
Pu Luo$^6$, Yumei Li$^6$, Cong Xu$^6$, Jinming Chai$^6$, Kexin Zhang$^6$, Licheng Jiao$^6$, Lingling Li$^6$ & 
$^6$ Xidian University, Xi'an, China & 25171214094@stu. xidian.edu.cn \\
\hline

ZYS & 
Siqi Yu$^7$, Chao Zhang$^7$, Kehuan Song$^7$, Fang Liu$^7$, Puhua Chen$^7$, Xu Liu$^7$ & 
$^7$ School of Artificial Intelligence, Xidian University, Xi'an, China & 24171213883@stu. xidian.edu.cn \\
\hline

Simplehh & 
Jin Hu$^8$, Jinyang Xu$^8$, Biao Liu$^8$ & 
$^8$ Xidian University, Xi'an, China & 18048853347@163. com \\
\hline

\end{tabular}
\caption{Teams, members, and affiliations for AIM 2025 Rip Current Segmentation Challenge.}
\label{tab:teams_affil}
\end{table*}

\subsection{Team RipEye}

\quad The team employed the HRDA unsupervised domain adaptation model \cite{hoyer2022daformer}, trained with the domain generalization strategy from SHADE \cite{zhao2022style}, to address the significant feature gap between the training and validation sets in the RipSeg challenge. This gap arises from the difficulty of constructing datasets that capture the full complexity and variability of rip currents. Domain generalization was selected for its ability to improve robustness to unseen conditions, such as variations in viewpoint and weather, thereby enhancing model generalization with limited training diversity. Since the pretraining dataset lacks rip current examples, the retrospection consistency module from SHADE was removed, as it relies on pretraining-based knowledge transfer to reduce overfitting. RipGAN \cite{qian2025ripgan} was also incorporated to generate synthetic rip current images from bounding boxes or polygons, increasing data diversity and providing SHADE with a wider range of styles for learning invariant features. The training architecture is shown in Figure \ref{fig:RipEye}.

\begin{figure*}[t]
    \centering
    \includegraphics[width=0.75\textwidth]{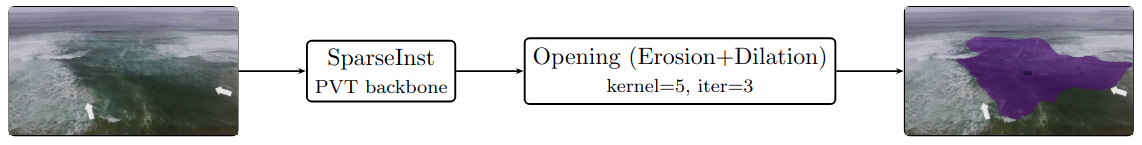}
    \caption{Overview of team RipSense's workflow showing the predicted segmentation mask of a random test sample, using SparseInst as the instance segmentation model, followed by morphological opening.} 
    \label{fig:aakash_pipeline}
\end{figure*}

\noindent
\textbf{Reproducibility details.} To train the proposed architecture, team RipEye employed the following settings:
\begin{itemize}
  \item The team used AdamW ($\beta_1=0.9,\beta_2=0.999$) as the optimizer, with a weight decay of $0.01$. The initial learning rate was $6\cdot 10^{-5}$ for the encoder and $6\cdot 10^{-4}$ for the decoder. The learning rate was linearly increased during warm-up for the first 1,500 iterations, followed by linear decay.
  \item In addition to using the training data provided by the organizers, the team also randomly selected polygons from 50\% of images with rip currents in the training set to generate synthetic data (5,141 images) for training.
  \item The team followed the HRDA training settings, which adopted random cropping ($1024\times1024$) and flipping. The size of the high-resolution detail cropping is $384\times384$. The number of basic styles is $98$. Additionally, the confidence score was obtained by subtracting the background probability from the rip current probability. During inference, they used only low-resolution context cropping to reduce computational cost, as it introduces no significant performance degradation.
  \item Training and inference were performed on an NVIDIA RTX 4090 with 24GB VRAM.
\end{itemize}

\begin{figure*}[t]
    \centering
    \includegraphics[width=0.75\textwidth]{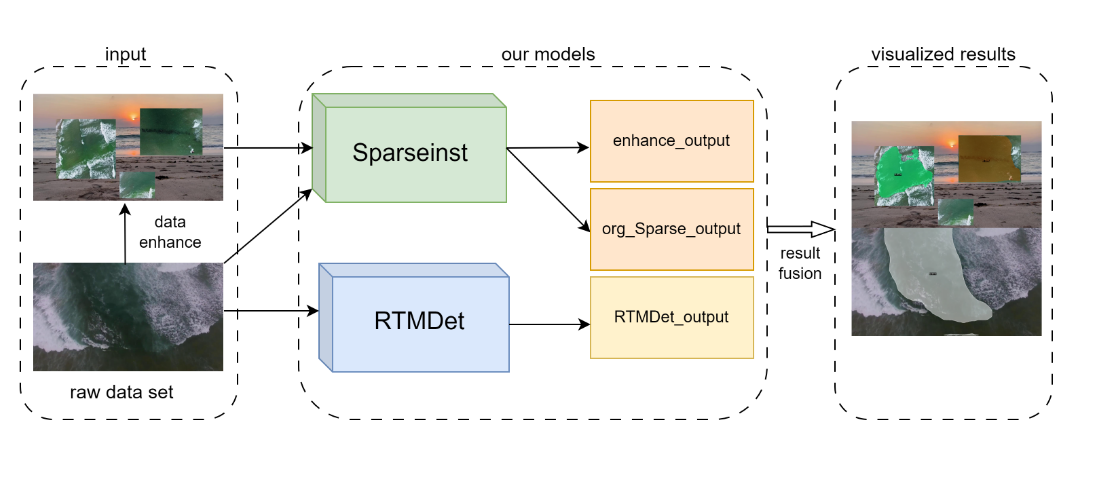}
    \vspace{-0.8cm}
    \caption{Rip current segmentation approach by team Gogogochufalou. Raw data is augmented, then processed by SparseInst (outputting enhanced and original results) and RTMDet. Outputs are fused to generate visualized rip current segmentation results.}
    \label{fig:gogochufalou}
\end{figure*}

\subsection{Team RipSense}

\quad The RipSense team proposed a pipeline that employs SparseInst \cite{cheng2022sparseinstanceactivationrealtime} with a PVTv2-B1 backbone, followed by a post-processing step of morphological opening (erosion, then dilation) to refine and smooth the segmentation masks of the rip currents, as illustrated in Figure \ref{fig:aakash_pipeline}. This step was particularly important as rip currents are amorphous and vary greatly in shape. 

\noindent
\textbf{Ablation and observations.} A range of real-time, state-of-the-art architectures were evaluated by the team, including CNN-based models (YOLO11 \cite{khanam2024yolov11} variants, SparseInst) and transformer-based models (RTMDet \cite{lyu2022rtmdetempiricalstudydesigning} variants), before choosing SparseInst for this specific task. Based on qualitative results on the validation dataset,
the following observations were made:
\begin{itemize}
    \item One key challenge with CNN-based approaches like YOLO11 was their tendency to produce multiple overlapping predictions for the same amorphous and irregularly shaped rip current. These overlaps often misaligned, making standard Non-Max-Suppression (NMS) thresholds (0.6–0.7) ineffective; lower thresholds (0.3-0.4) sometimes helped, but also removed true positives when multiple rip currents were close together. Transformer-based models like RTMDet and the unique case of SparseInst, which, despite being fully convolutional, uses bipartite matching and instance activation maps for one-to-one prediction, avoided this issue by eliminating duplicates without relying heavily on NMS.
    \item CNN-based models, due to their inductive bias, occasionally outperformed transformers in cases where local texture cues, such as foam or sharp wave patterns, made rip currents distinct from surrounding water.
    \item In contrast, transformer-based models performed better when a broader spatial context was essential, such as in scenes where rip currents blended into the background and could only be detected by considering global context across the image, for example in the case of sediment rip currents.
\end{itemize}

\noindent
\textbf{Reproducibility details.} The following settings were used to configure the model:
\begin{itemize}
    \item The SparseInst model was trained for 10~epochs using AdamW (learning rate $5\cdot 10^{-5}$, $\beta_{1}=0.9$, $\beta_{2}=0.999$, $\epsilon=1\cdot 10^{-8}$, weight decay $0.05$) with a WarmupMultiStepLR schedule (linear warmup for the first $400$ iterations at factor $0.001$; decays at $6$ and $8$ epochs with a factor of $0.1$), a batch size of $4$ on an NVIDIA RTX 3060 GPU, and horizontal flip augmentation.
    \item During inference, a confidence score threshold of $0.4$ was used to consider a prediction as valid.
    \item The post-processing step consisted of morphological opening with a kernel size of $5$ for $3$ iterations.
\end{itemize}

\begin{figure*}[t]
    \centering
    \includegraphics[width=0.75\textwidth]{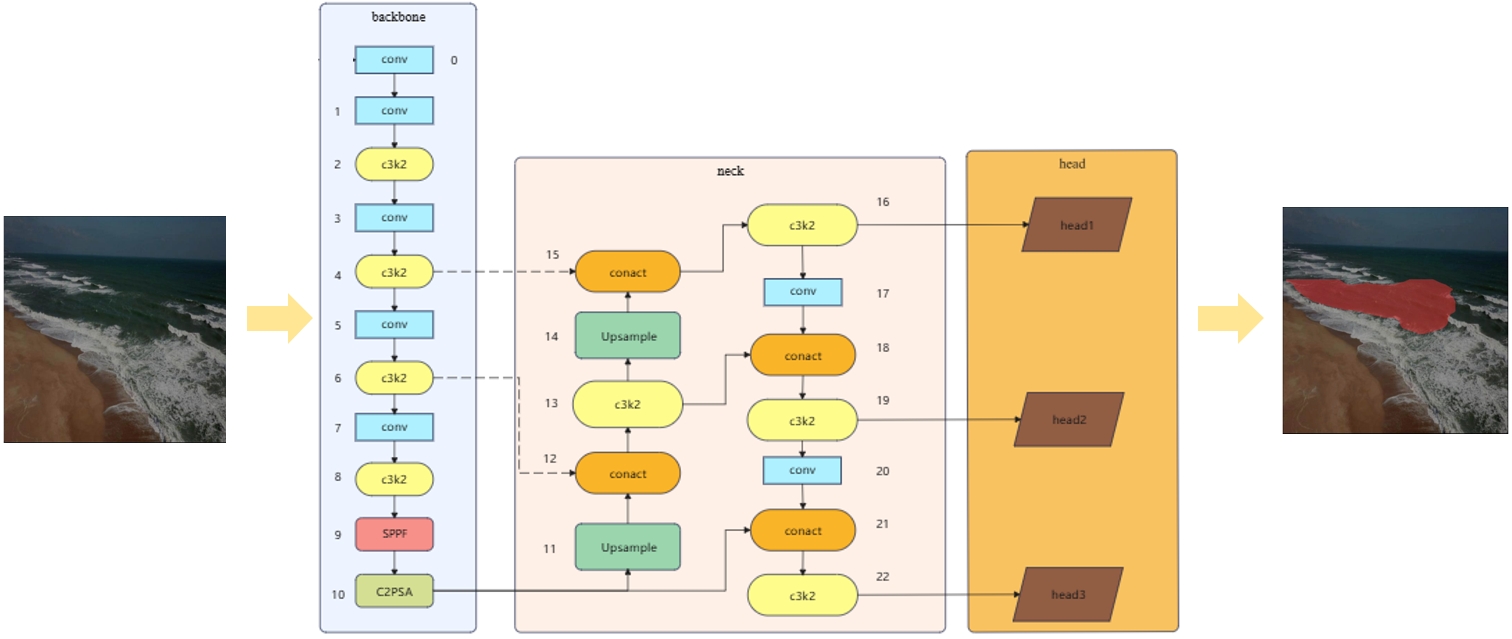}
    \caption{Team ZYS's scheme for rip current segmentation, using YOLO11x.}
    \label{fig:zys_team}
\end{figure*}

\begin{figure*}[t]
    \centering
    \includegraphics[width=0.75\textwidth]{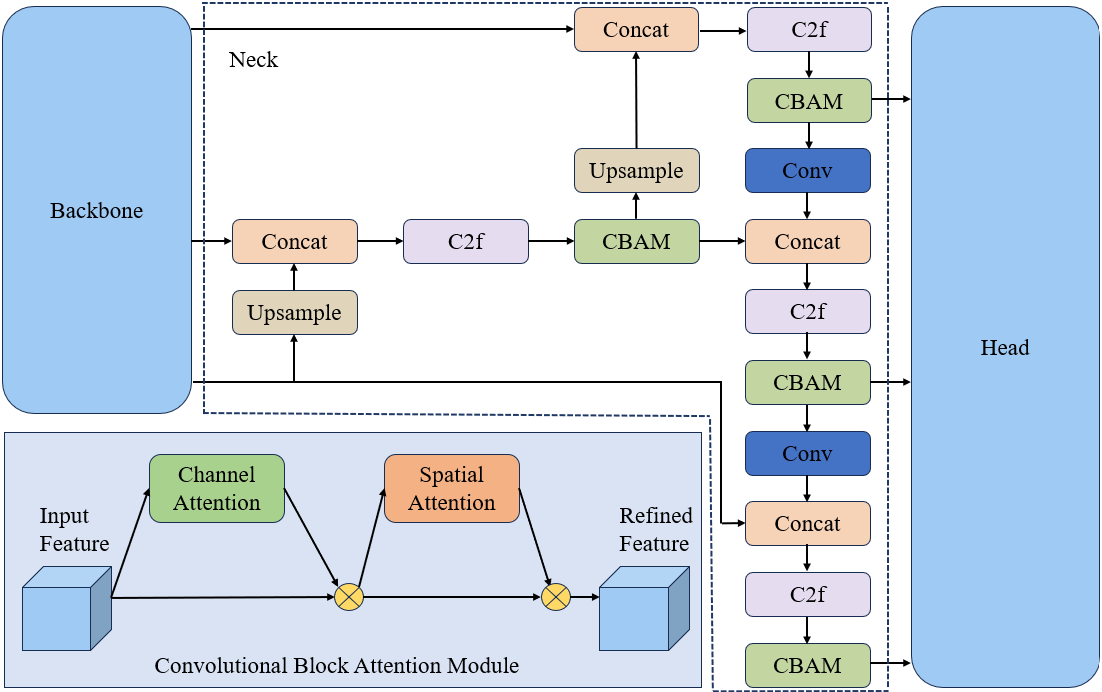}
    \caption{Team Simplehh's fine-tuned YOLOv8n Pipeline. Four CBAM blocks were inserted after the C2f block in the Neck part to enhance the model's ability to discriminate similar features between foreground and background.}
    \label{fig:Simplehh}
\end{figure*}

\subsection{Team Gogogochufalou} 

\quad The team proposed a dual-model collaborative framework for rip current segmentation, integrating SparseInst \cite{cheng2022sparseinstanceactivationrealtime} and RTMDet \cite{lyu2022rtmdetempiricalstudydesigning} to leverage their complementary strengths.
In this approach, a trained base model predicted unlabeled data, with low-confidence samples used as blank backgrounds for pasting rip current instances from the training set, limited to a maximum of three per generated image. SparseInst was configured with a PVTv2-B2-li backbone, a single target class, and 150 masks, while RTMDet was initialized from RTMDet-Ins-x pre-trained weights and fine-tuned for the task. Prediction fusion was performed using IoU and Dice coefficient thresholds to identify matching detections, with the higher-confidence result retained. Figure~\ref{fig:gogochufalou} illustrates the overall pipeline of this rip current segmentation approach, encompassing data processing, model training, and result fusion stages.

\noindent
\textbf{Reproducibility details.} The following configurations were used during model training and validation:
\begin{itemize}
    \item SparseInst training used a base learning rate of $0.0000125$, a batch size of $4$, and a maximum image resolution of $1333 \times 1333$. RTMDet retained its original architecture with end-to-end fine-tuning.
    \item Data augmentation techniques included rotation, scaling, flipping, and color jittering, along with advanced augmentation via the Simple Copy-Paste method \cite{ghiasi2021simplecopypastestrongdata}.
    \item A multi-scale training strategy was used for RTMDet, enhancing the model's adaptability to different-sized rip currents by randomly adjusting the input image scale.
\item Both models were trained on a heterogeneous GPU cluster: Node A with 4 $\times$ NVIDIA GeForce RTX 2080 Ti GPUs, and Node B with 2 $\times$ NVIDIA GeForce RTX 3090 GPUs.
\end{itemize}

\subsection{Team ZYS}

\quad Team ZYS selected the YOLO11x model \cite{khanam2024yolov11} for the RipSeg Challenge. It was chosen due to its exceptional performance in the field of instance segmentation. The architecture of the method is shown in Figure~\ref{fig:zys_team}. It adopts a deeper network structure and an optimized feature fusion mechanism, enabling effective capture of rip current features at various scales in images. Compared with traditional segmentation models, this model, relying on its end-to-end detection architecture, can significantly improve processing speed while ensuring high-precision segmentation, which is crucial for scenarios with high timeliness requirements such as real-time rip current monitoring \cite{arun2014rip,desilva2021automated,dudkowska2020rip,hong2021numerical}.

\noindent
\textbf{Reproducibility details.} The team used the followings settings to configure their model:
\begin{itemize}
    \item The team trained the model for 45 epochs. To ensure sufficient convergence of the model on the RipSeg dataset, a dynamic learning rate adjustment strategy was employed during training. The initial learning rate was set to $0.001$, and it decayed to $0.1$ times the original value every $10$ epochs as the training progressed, thereby balancing the convergence speed and stability of the model.
    \item Meanwhile, to enhance the generalization ability of the model, the team performed multi-dimensional data augmentation operations on the training data, including random horizontal flipping (with a probability of $0.5$), brightness and contrast adjustment (within a range of $\pm0.2$), and random cropping (with a size range of $0.7$-$1.0$ times the original image), which effectively alleviated the potential distribution bias of the dataset.
    \item The key hyperparameter settings in the inference phase were optimized through multiple sets of comparative experiments. Among them, the confidence threshold was set to $0.15$, which was determined after comprehensively considering the fuzzy characteristics of rip current regions. A lower confidence threshold was used to retain more potential rip current regions, prioritizing the reduction of false negatives given the safety-critical nature of the task. The Intersection over Union (IoU) threshold for NMS was set to $0.6$. The maximum number of detections was set to $300$.
\end{itemize}

\subsection{Team Simplehh}

\quad The team chose to fine-tune YOLOv8n model \cite{sohan2024reviewyolov8} on the RipSeg dataset, after comparing with the baseline
in RipVIS \cite{Dumitriu_2025_CVPR}. The competition’s data contains video frames, which exhibit minimal differences between adjacent frames and high similarity between foreground and background features. As a lightweight model, YOLOv8n reduces the risk of overfitting to noise or subtle fluctuations in such data, while enabling rapid convergence to core patterns despite the redundancy. To address the limited feature discriminability of lightweight architectures, several CBAM \cite{woo2018cbamconvolutionalblockattention} blocks were incorporated into the Neck (Figure \ref{fig:Simplehh}), enhancing the emphasis on key foreground channels and suppressing background interference, without significantly increasing parameters or inference latency.

\noindent
\textbf{Reproducibility details.} The configurations steps are as follows:
\begin{itemize}
\item The model was trained using the AdamW optimizer with $\beta_1=0.9$ and $\beta_2=0.999$, where the learning rate gradually decreased from $0.01$ to $0.001$. The image size was $640\times640$ and and the batch size was $64$. The model underwent $50$ epochs of training on the original training dataset. The training objectives were based on binary cross-entropy loss, complete IoU loss, dice loss, and cross-entropy loss.
\item Data augmentation methods included HSV adjustment, random cropping, image translation, horizontal flipping, random erasing, and mix-up.
\item The team carried out the experiments using a single vGPU-32 device provided by the AutoDL platform, with PyTorch 2.7.1. 
\end{itemize}

\section{Conclusion and Future Work}
This report presented the AIM 2025 Rip Current Segmentation (RipSeg) Challenge and provided a comprehensive ranking of five participating frameworks, evaluated across benchmark metrics such as $F_1$, $F_2$, $AP_{50}$, and $AP_{[50:95]}$, in the context of instance segmentation. The submitted solutions showcased a diverse range of strategies, spanning lightweight CNNs, transformer-based architectures, domain adaptation techniques, synthetic data generation, and morphological post-processing. Approaches centered on domain generalization, tailored model design and data augmentation demonstrated particular promise, highlighting the multiple research avenues available for tackling this safety-critical task.

The final leaderboard, summarized in Table~\ref{tab:test_results}, underscores the difficulty of rip current segmentation: even the top-performing method achieved a computed score of only $0.68$. These results indicate that existing state-of-the-art models, whether directly applied or adapted through domain transfer, are not yet sufficient for robust rip current detection. This points to the need for novel methodological advances that go beyond current paradigms.

Looking forward, future editions of RipSeg can explore expanded architectures, integration of temporal information from video, the use of multi-modal data sources or other contextual cues that influence rip current formation. Additionally, broader participation and shared open baselines may accelerate progress in this domain. By continuing to refine both the dataset and evaluation framework, the RipSeg challenge aims to catalyze the development of more accurate, reliable, and deployable solutions for real-world beach safety applications.

\section*{Acknowledgments}
This work was partially supported by the Alexander von Humboldt Foundation. We thank the University of Bucharest and the AIM 2025 sponsors: AI Witchlabs and University of W\"urzburg (Computer Vision Lab).

{
    \small
    \bibliographystyle{ieeenat_fullname}
    \bibliography{main}

\begin{thebibliography}{58}
\providecommand{\natexlab}[1]{#1}
\providecommand{\url}[1]{\texttt{#1}}
\expandafter\ifx\csname urlstyle\endcsname\relax
  \providecommand{\doi}[1]{doi: #1}\else
  \providecommand{\doi}{doi: \begingroup \urlstyle{rm}\Url}\fi

\bibitem[Arun~Kumar and Prasad(2014)]{arun2014rip}
S.V.V. Arun~Kumar and K.V.S.R. Prasad.
\newblock {Rip current-related fatalities in India: a new predictive risk scale for forecasting rip currents}.
\newblock \emph{Natural Hazards}, 70:\penalty0 313--335, 2014.

\bibitem[Brander et~al.(2013)Brander, Dominey-Howes, Champion, Del~Vecchio, and Brighton]{brander2013brief}
R. Brander, Dale Dominey-Howes, C. Champion, O. Del~Vecchio, and B. Brighton.
\newblock {Brief Communication: A new perspective on the Australian rip current hazard}.
\newblock \emph{Natural Hazards and Earth System Sciences}, 13\penalty0 (6):\penalty0 1687--1690, 2013.

\bibitem[Brander and Short(2000)]{brander2000morphodynamics}
Robert~W. Brander and A.D. Short.
\newblock {Morphodynamics of a large-scale rip current system at Muriwai Beach, New Zealand}.
\newblock \emph{Marine Geology}, 165\penalty0 (1-4):\penalty0 27--39, 2000.

\bibitem[Brewster et~al.(2019)Brewster, Gould, and Brander]{brewster2019estimations}
B.~Chris Brewster, Richard~E. Gould, and Robert~W. Brander.
\newblock {Estimations of rip current rescues and drowning in the United States}.
\newblock \emph{Natural Hazards and Earth System Sciences}, 19\penalty0 (2):\penalty0 389--397, 2019.

\bibitem[Castelle et~al.(2016)Castelle, Scott, Brander, and McCarroll]{castelle2016rip}
B. Castelle, Tim Scott, R.W. Brander, and R.J. McCarroll.
\newblock Rip current types, circulation and hazard.
\newblock \emph{Earth-Science Reviews}, 163:\penalty0 1--21, 2016.

\bibitem[Cheng et~al.(2022)Cheng, Wang, Chen, Zhang, Zhang, Huang, Zhang, and Liu]{cheng2022sparseinstanceactivationrealtime}
Tianheng Cheng, Xinggang Wang, Shaoyu Chen, Wenqiang Zhang, Qian Zhang, Chang Huang, Zhaoxiang Zhang, and Wenyu Liu.
\newblock Sparse instance activation for real-time instance segmentation, 2022.

\bibitem[Choi et~al.(2024)Choi, Rajendran, and Suh]{choi2024explainable}
Juno Choi, Muralidharan Rajendran, and Yong~Cheol Suh.
\newblock {Explainable Rip Current Detection and Visualization with XAI EigenCAM}.
\newblock In \emph{Proceedings of 26th International Conference on Advanced Communications Technology}, pages 1--6, 2024.

\bibitem[Ciubotariu et~al.(2025)Ciubotariu, Vasluianu, Zhou, Mehta, Timofte, et~al.]{aim2025highfps}
George Ciubotariu, Florin-Alexandru Vasluianu, Zhuyun Zhou, Nancy Mehta, Radu Timofte, et~al.
\newblock {AIM} 2025 high {FPS} non-uniform motion deblurring challenge report.
\newblock In \emph{Proceedings of the IEEE/CVF International Conference on Computer Vision (ICCV) Workshops}, 2025.

\bibitem[Cordts et~al.(2016)Cordts, Omran, Ramos, Rehfeld, Enzweiler, Benenson, Franke, Roth, and Schiele]{cordts2016cityscapes}
Marius Cordts, Mohamed Omran, Sebastian Ramos, Timo Rehfeld, Markus Enzweiler, Rodrigo Benenson, Uwe Franke, Stefan Roth, and Bernt Schiele.
\newblock The cityscapes dataset for semantic urban scene understanding.
\newblock In \emph{Proceedings of the IEEE Conference on Computer Vision and Pattern Recognition (CVPR)}, pages 3213--3223, 2016.

\bibitem[Da~F.~Klein et~al.(2003)Da~F.~Klein, Santana, Diehl, and De~Menezes]{da2003analysis}
A.H. Da~F.~Klein, G.G. Santana, F.L. Diehl, and J.T. De~Menezes.
\newblock {Analysis of hazards associated with sea bathing: results of five years work in oceanic beaches of Santa Catarina state, southern Brazil}.
\newblock \emph{Journal of Coastal Research}, pages 107--116, 2003.

\bibitem[de~Silva et~al.(2021{\natexlab{a}})de~Silva, Mori, Dusek, Davis, and Pang]{desilva2021automated}
Akila de Silva, Issei Mori, Gregory Dusek, James Davis, and Alex Pang.
\newblock Automated rip current detection with region based convolutional neural networks.
\newblock \emph{Coastal Engineering}, 166:\penalty0 103859, 2021{\natexlab{a}}.
\newblock 2, 3.

\bibitem[de~Silva et~al.(2021{\natexlab{b}})de~Silva, Mori, Dusek, Davis, and Pang]{desilva2021frcnn}
Akila de Silva, Issei Mori, Gregory Dusek, James Davis, and Alex Pang.
\newblock Automated rip current detection with region based convolutional neural networks.
\newblock \emph{Coastal Engineering}, 166:\penalty0 103859, 2021{\natexlab{b}}.

\bibitem[de~Silva et~al.(2024)de~Silva, Zhao, Stewart, Hasan, Dusek, Davis, and Pang]{de2023ripviz}
Akila de Silva, Mona Zhao, Donald Stewart, Fahim Hasan, Gregory Dusek, James Davis, and Alex Pang.
\newblock {RipViz: Finding Rip Currents by Learning Pathline Behavior}.
\newblock \emph{IEEE Transactions on Visualization and Computer Graphics}, 30\penalty0 (7):\penalty0 3930--3944, 2024.

\bibitem[Dudkowska et~al.(2020)Dudkowska, Borun, Malicki, Sch{\"o}nhofer, and Gic-Grusza]{dudkowska2020rip}
Aleksandra Dudkowska, Aleksandra Borun, Jakub Malicki, Jan Sch{\"o}nhofer, and Gabriela Gic-Grusza.
\newblock Rip currents in the non-tidal surf zone with sandbars: numerical analysis versus field measurements.
\newblock \emph{Oceanologia}, 62\penalty0 (3):\penalty0 291--308, 2020.

\bibitem[Dumitriu et~al.(2023)Dumitriu, Tatui, Miron, Ionescu, and Timofte]{dumitriu2023rip}
Andrei Dumitriu, Florin Tatui, Florin Miron, Radu~Tudor Ionescu, and Radu Timofte.
\newblock {Rip Current Segmentation: A novel benchmark and YOLOv8 baseline results}.
\newblock In \emph{Proceedings of the IEEE/CVF Conference on Computer Vision and Pattern Recognition (CVPR) Workshops}, pages 1261--1271, 2023.

\bibitem[Dumitriu et~al.(2025{\natexlab{a}})Dumitriu, Miron, Tatui, Ionescu, Timofte, Ralhan, Vasluianu, et~al.]{aim2025ripseg}
Andrei Dumitriu, Florin Miron, Florin Tatui, Radu~Tudor Ionescu, Radu Timofte, Aakash Ralhan, Florin-Alexandru Vasluianu, et~al.
\newblock {AIM} 2025 challenge on rip current segmentation ({RipSeg}).
\newblock In \emph{Proceedings of the IEEE/CVF International Conference on Computer Vision (ICCV) Workshops}, 2025{\natexlab{a}}.

\bibitem[Dumitriu et~al.(2025{\natexlab{b}})Dumitriu, Tatui, Miron, Ralhan, Ionescu, and Timofte]{Dumitriu_2025_CVPR}
Andrei Dumitriu, Florin Tatui, Florin Miron, Aakash Ralhan, Radu~Tudor Ionescu, and Radu Timofte.
\newblock Ripvis: Rip currents video instance segmentation benchmark for beach monitoring and safety.
\newblock In \emph{Proceedings of the IEEE/CVF Conference on Computer Vision and Pattern Recognition (CVPR)}, pages 3427--3437, 2025{\natexlab{b}}.

\bibitem[Feijoo et~al.(2025)Feijoo, Garrido, Conde, Rim, Garcia, Cho, Timofte, et~al.]{aim2025efficientdeblurring}
Daniel Feijoo, Paula Garrido, Marcos Conde, Jaesung Rim, Alvaro Garcia, Sunghyun Cho, Radu Timofte, et~al.
\newblock Efficient real-world deblurring using single images: {AIM} 2025 challenge report.
\newblock In \emph{Proceedings of the IEEE/CVF International Conference on Computer Vision (ICCV) Workshops}, 2025.

\bibitem[Ghiasi et~al.(2021)Ghiasi, Cui, Srinivas, Qian, Lin, Cubuk, Le, and Zoph]{ghiasi2021simplecopypastestrongdata}
Golnaz Ghiasi, Yin Cui, Aravind Srinivas, Rui Qian, Tsung-Yi Lin, Ekin~D. Cubuk, Quoc~V. Le, and Barret Zoph.
\newblock Simple copy-paste is a strong data augmentation method for instance segmentation.
\newblock In \emph{Proceedings of the IEEE/CVF Conference on Computer Vision and Pattern Recognition (CVPR)}, pages 2918--2928, 2021.

\bibitem[He et~al.(2017)He, Gkioxari, Doll{\'a}r, and Girshick]{he2017mask}
Kaiming He, Georgia Gkioxari, Piotr Doll{\'a}r, and Ross Girshick.
\newblock {Mask R-CNN}.
\newblock In \emph{Proceedings of the IEEE International Conference on Computer Vision (ICCV)}, pages 2961--2969, 2017.

\bibitem[Hong et~al.(2021)Hong, Zhang, Wang, Zhou, Yu, and Zhang]{hong2021numerical}
Xiao Hong, Yao Zhang, Bin Wang, Shuihua Zhou, Shengbin Yu, and Juan Zhang.
\newblock Numerical study of rip currents interlaced with multichannel sandbars.
\newblock \emph{Natural Hazards}, 108\penalty0 (1):\penalty0 593--605, 2021.

\bibitem[Hoyer et~al.(2022)Hoyer, Dai, and Van~Gool]{hoyer2022daformer}
Lars Hoyer, Dengxin Dai, and Luc Van~Gool.
\newblock Daformer: Improving network architectures and training strategies for domain-adaptive semantic segmentation.
\newblock In \emph{Proceedings of the IEEE/CVF Conference on Computer Vision and Pattern Recognition (CVPR)}, pages 9924--9935, 2022.

\bibitem[Ignatov et~al.(2025{\natexlab{a}})Ignatov, Perevozchikov, Timofte, et~al.]{aim20254ksr}
Andrey Ignatov, Georgy Perevozchikov, Radu Timofte, et~al.
\newblock {4K} image super-resolution on mobile {NPUs}: {Mobile AI \& AIM 2025} challenge report.
\newblock In \emph{Proceedings of the IEEE/CVF International Conference on Computer Vision (ICCV) Workshops}, 2025{\natexlab{a}}.

\bibitem[Ignatov et~al.(2025{\natexlab{b}})Ignatov, Perevozchikov, Timofte, et~al.]{aim2025efficientISP}
Andrey Ignatov, Georgy Perevozchikov, Radu Timofte, et~al.
\newblock Efficient learned smartphone {ISP} on mobile {GPUs}: {Mobile AI \& AIM 2025} challenge report.
\newblock In \emph{Proceedings of the IEEE/CVF International Conference on Computer Vision (ICCV) Workshops}, 2025{\natexlab{b}}.

\bibitem[Ignatov et~al.(2025{\natexlab{c}})Ignatov, Perevozchikov, Timofte, et~al.]{aim2025efficientdenoising}
Andrey Ignatov, Georgy Perevozchikov, Radu Timofte, et~al.
\newblock Efficient image denoising on smartphone {GPUs}: {Mobile AI \& AIM 2025} challenge report.
\newblock In \emph{Proceedings of the IEEE/CVF International Conference on Computer Vision (ICCV) Workshops}, 2025{\natexlab{c}}.

\bibitem[Ignatov et~al.(2025{\natexlab{d}})Ignatov, Perevozchikov, Timofte, et~al.]{aim2025sd}
Andrey Ignatov, Georgy Perevozchikov, Radu Timofte, et~al.
\newblock Adapting stable diffusion for on-device inference: {Mobile AI \& AIM 2025} challenge report.
\newblock In \emph{Proceedings of the IEEE/CVF International Conference on Computer Vision (ICCV) Workshops}, 2025{\natexlab{d}}.

\bibitem[Jocher et~al.(2023)Jocher, Chaurasia, and Qiu]{Jocher_YOLO_by_Ultralytics_2023}
Glenn Jocher, Ayush Chaurasia, and Jing Qiu.
\newblock {YOLO by Ultralytics}, 2023.

\bibitem[Karetin et~al.(2025)Karetin, Molodetskikh, Vatolin, Timofte, et~al.]{aim2025videoSR}
Nikolai Karetin, Ivan Molodetskikh, Dmitry Vatolin, Radu Timofte, et~al.
\newblock {AIM} 2025 challenge on robust offline video super-resolution: Dataset, methods and results.
\newblock In \emph{Proceedings of the IEEE/CVF International Conference on Computer Vision (ICCV) Workshops}, 2025.

\bibitem[Khan et~al.(2025{\natexlab{a}})Khan, De~Silva, Palinkas, Dusek, Davis, and Pang]{khan2025ripfinder}
Fahim Khan, Akila De~Silva, Ashleigh Palinkas, Gregory Dusek, James Davis, and Alex Pang.
\newblock {RipFinder: Real-time rip current detection on mobile devices}.
\newblock \emph{Frontiers in Marine Science}, 12:\penalty0 1549513, 2025{\natexlab{a}}.

\bibitem[Khan et~al.(2025{\natexlab{b}})Khan, Stewart, de~Silva, Palinkas, Dusek, Davis, and Pang]{khan2025ripscout}
Fahim Khan, Donald Stewart, Akila de Silva, Ashleigh Palinkas, Gregory Dusek, James Davis, and Alex Pang.
\newblock Ripscout: Realtime ml-assisted rip current detection and automated data collection using uavs.
\newblock \emph{IEEE Journal of Selected Topics in Applied Earth Observations and Remote Sensing}, 2025{\natexlab{b}}.

\bibitem[Khanam and Hussain(2024)]{khanam2024yolov11}
R Khanam and M Hussain.
\newblock {YOLOv11}: An overview of the key architectural enhancements.
\newblock \emph{arXiv preprint arXiv:2410.17725}, 2024.
\newblock 2.

\bibitem[Kirillov et~al.(2023)Kirillov, Mintun, Ravi, Mao, Rolland, Gustafson, Xiao, Whitehead, Berg, Lo, et~al.]{kirillov2023segment}
Alexander Kirillov, Eric Mintun, Nikhila Ravi, Hanzi Mao, Chloe Rolland, Laura Gustafson, Tete Xiao, Spencer Whitehead, Alexander~C Berg, Wan-Yen Lo, et~al.
\newblock Segment anything.
\newblock In \emph{Proceedings of the IEEE/CVF International Conference on Computer Vision (ICCV)}, pages 4015--4026, 2023.

\bibitem[Li et~al.(2025)Li, Li, Conde, Besbinar, Hosu, Iso, Timofte, et~al.]{aim2025rawdenoising}
Feiran Li, Jiacheng Li, Marcos Conde, Beril Besbinar, Vlad Hosu, Daisuke Iso, Radu Timofte, et~al.
\newblock Real-world raw denoising using diverse cameras: {AIM} 2025 challenge report.
\newblock In \emph{Proceedings of the IEEE/CVF International Conference on Computer Vision (ICCV) Workshops}, 2025.

\bibitem[Lin et~al.(2014)Lin, Maire, Belongie, Hays, Perona, Ramanan, Doll{\'a}r, and Zitnick]{lin2014microsoft}
Tsung-Yi Lin, Michael Maire, Serge Belongie, James Hays, Pietro Perona, Deva Ramanan, Piotr Doll{\'a}r, and C~Lawrence Zitnick.
\newblock {Microsoft COCO: Common objects in context}.
\newblock In \emph{Proceedings of 13th European conference on Computer Vision (ECCV)}, pages 740--755. Springer, 2014.

\bibitem[Longarela et~al.(2025)Longarela, Conde, Álvaro García, Timofte, et~al.]{aim2025perceptual}
Bruno Longarela, Marcos Conde, Álvaro García, Radu Timofte, et~al.
\newblock {AIM} 2025 perceptual image super-resolution challenge.
\newblock In \emph{Proceedings of the IEEE/CVF International Conference on Computer Vision (ICCV) Workshops}, 2025.

\bibitem[Lushine(1991)]{lushine1991study}
James~B. Lushine.
\newblock A study of rip current drownings and related weather factors.
\newblock \emph{National Weather Digest}, 16\penalty0 (3):\penalty0 13--19, 1991.

\bibitem[Lyu et~al.(2022)Lyu, Zhang, Huang, Zhou, Wang, Liu, Zhang, and Chen]{lyu2022rtmdetempiricalstudydesigning}
Chengqi Lyu, Wenwei Zhang, Haian Huang, Yue Zhou, Yudong Wang, Yanyi Liu, Shilong Zhang, and Kai Chen.
\newblock {RTMDet: An Empirical Study of Designing Real-Time Object Detectors}.
\newblock \emph{arXiv preprint arXiv:2212.07784}, 2022.

\bibitem[Maryan et~al.(2019)Maryan, Hoque, Michael, Ioup, and Abdelguerfi]{maryan2019machine}
Corey Maryan, Md~Tamjidul Hoque, Christopher Michael, Elias Ioup, and Mahdi Abdelguerfi.
\newblock Machine learning applications in detecting rip channels from images.
\newblock \emph{Applied Soft Computing}, 78:\penalty0 84--93, 2019.

\bibitem[McGill and Ellis(2022)]{mcgill2022flow}
Sean~P. McGill and Jean~T. Ellis.
\newblock Rip current and channel detection using surfcams and optical flow.
\newblock \emph{Shore \& Beach}, 90\penalty0 (1):\penalty0 50, 2022.

\bibitem[Mori et~al.(2022)Mori, de~Silva, Dusek, Davis, and Pang]{mori2022flow}
Issei Mori, Akila de Silva, Gregory Dusek, James Davis, and Alex Pang.
\newblock Flow-based rip current detection and visualization.
\newblock \emph{IEEE Access}, 10:\penalty0 6483--6495, 2022.

\bibitem[{National Oceanic and Atmospheric Administration (NOAA)}(2023)]{noaa2023ripcurrents}
{National Oceanic and Atmospheric Administration (NOAA)}.
\newblock What is a rip current?
\newblock https://oceanservice.noaa.gov/facts/ripcurrent.html, 2023.
\newblock Accessed: March, 2023.

\bibitem[Philip and Pang(2016)]{philip2016flow}
Shweta Philip and Alex Pang.
\newblock {Detecting and Visualizing Rip Current Using Optical Flow}.
\newblock In \emph{Proceedings of the Eurographics / IEEE VGTC Conference on Visualization: Short Papers}, pages 19--23, 2016.

\bibitem[Qian et~al.(2025)Qian, Harley, Razzak, and Song]{qian2025ripgan}
Shenyang Qian, Mitchell Harley, Imran Razzak, and Yang Song.
\newblock {RipGAN: A GAN-based rip current data augmentation method}.
\newblock In \emph{Proceedings of the IEEE International Conference on Robotics and Automation}, 2025.

\bibitem[Rampal et~al.(2022)Rampal, Shand, Wooler, and Rautenbach]{rampal2022interpretable}
Neelesh Rampal, Tom Shand, Adam Wooler, and Christo Rautenbach.
\newblock Interpretable deep learning applied to rip current detection and localization.
\newblock \emph{Remote Sensing}, 14\penalty0 (23):\penalty0 6048, 2022.

\bibitem[Rashid et~al.(2020)Rashid, Razzak, Tanveer, and Robles-Kelly]{rashid2020ripnet}
Ashraf~Haroon Rashid, Imran Razzak, Muhammad Tanveer, and Antonio Robles-Kelly.
\newblock Ripnet: A lightweight one-class deep neural network for the identification of rip currents.
\newblock In \emph{Proceedings of 27th International Conference on Neural Information Processing}, pages 172--179, 2020.

\bibitem[Rashid et~al.(2021)Rashid, Razzak, Tanveer, and Robles-Kelly]{rashid2021ripdet}
Ashraf~Haroon Rashid, Imran Razzak, Muhammad Tanveer, and Antonio Robles-Kelly.
\newblock {RipDet: A fast and lightweight deep neural network for rip currents detection}.
\newblock In \emph{Proceedings of 2021 International Joint Conference on Neural Networks}, pages 1--6, 2021.

\bibitem[Rashid et~al.(2023)Rashid, Razzak, Tanveer, and Hobbs]{rashid2023reducing}
Ashraf~Haroon Rashid, Imran Razzak, M. Tanveer, and Michael Hobbs.
\newblock Reducing rip current drowning: An improved residual based lightweight deep architecture for rip detection.
\newblock \emph{ISA Transactions}, 132:\penalty0 199--207, 2023.

\bibitem[Ravi et~al.(2024)Ravi, Gabeur, Hu, Hu, Ryali, Ma, Khedr, R{\"a}dle, Rolland, Gustafson, et~al.]{ravi2024sam}
Nikhila Ravi, Valentin Gabeur, Yuan-Ting Hu, Ronghang Hu, Chaitanya Ryali, Tengyu Ma, Haitham Khedr, Roman R{\"a}dle, Chloe Rolland, Laura Gustafson, et~al.
\newblock {SAM 2: Segment Anything in Images and Videos}.
\newblock \emph{arXiv preprint arXiv:2408.00714}, 2024.

\bibitem[Safonov et~al.(2025)Safonov, Rakhmanov, Vatolin, Timofte, et~al.]{aim2025scvqa}
Nickolay Safonov, Mikhail Rakhmanov, Dmitriy Vatolin, Radu Timofte, et~al.
\newblock {AIM} 2025 challenge on screen-content video quality assessment: Methods and results.
\newblock In \emph{Proceedings of the IEEE/CVF International Conference on Computer Vision (ICCV) Workshops}, 2025.

\bibitem[Sohan et~al.(2024)Sohan, Sai~Ram, and Rami~Reddy]{sohan2024reviewyolov8}
Mupparaju Sohan, Thotakura Sai~Ram, and Ch~Venkata Rami~Reddy.
\newblock {A review on YOLOv8 and its advancements}.
\newblock In \emph{Proceedings of the International Conference on Data Intelligence and Cognitive Informatics}, pages 529--545. Springer, 2024.

\bibitem[Wang et~al.(2025)Wang, Banterle, Ren, Timofte, et~al.]{aim2025tone}
Chao Wang, Francesco Banterle, Bin Ren, Radu Timofte, et~al.
\newblock {AIM} 2025 challenge on inverse tone mapping report: Methods and results.
\newblock In \emph{Proceedings of the IEEE/CVF International Conference on Computer Vision (ICCV) Workshops}, 2025.

\bibitem[Woo et~al.(2018)Woo, Park, Lee, and Kweon]{woo2018cbamconvolutionalblockattention}
Sanghyun Woo, Jongchan Park, Joon-Young Lee, and In~So Kweon.
\newblock {CBAM: Convolutional Block Attention Module}.
\newblock In \emph{Proceedings of the European Conference on Computer Vision (ECCV)}, pages 3--19, 2018.

\bibitem[Xu et~al.(2022)Xu, Escalera, Pavão, Richard, Tu, Yao, Zhao, and Guyon]{codabench}
Zhen Xu, Sergio Escalera, Adrien Pavão, Magali Richard, Wei-Wei Tu, Quanming Yao, Huan Zhao, and Isabelle Guyon.
\newblock Codabench: Flexible, easy-to-use, and reproducible meta-benchmark platform.
\newblock \emph{Patterns}, 3\penalty0 (7):\penalty0 100543, 2022.

\bibitem[Yakovenko et~al.(2025)Yakovenko, Chakvetadze, Khrapov, Zhelezov, Vatolin, Timofte, et~al.]{aim2025videodenoising}
Alexander Yakovenko, George Chakvetadze, Ilya Khrapov, Maksim Zhelezov, Dmitry Vatolin, Radu Timofte, et~al.
\newblock {AIM} 2025 low-light raw video denoising challenge: Dataset, methods and results.
\newblock In \emph{Proceedings of the IEEE/CVF International Conference on Computer Vision (ICCV) Workshops}, 2025.

\bibitem[Yang et~al.(2019)Yang, Fan, and Xu]{Yang2019vis}
Linjie Yang, Yuchen Fan, and Ning Xu.
\newblock Video instance segmentation.
\newblock In \emph{Proceedings of the IEEE/CVF International Conference on Computer Vision (ICCV)}, pages 5188--5197, 2019.

\bibitem[Yang et~al.(2021)Yang, Fan, Fu, and Xu]{vis2021}
Linjie Yang, Yuchen Fan, Yang Fu, and Ning Xu.
\newblock The 3rd large-scale video object segmentation challenge - video instance segmentation track, 2021.

\bibitem[Zhao et~al.(2022)Zhao, Zhong, Zhao, Sebe, and Lee]{zhao2022style}
Yuyuan Zhao, Zaiyu Zhong, Nan Zhao, Nicu Sebe, and Gim~Hee Lee.
\newblock Style-hallucinated dual consistency learning for domain generalized semantic segmentation.
\newblock In \emph{Proceedings of the European Conference on Computer Vision (ECCV)}, pages 535--552, 2022.

\bibitem[Zhu et~al.(2022)Zhu, Qi, Hu, Su, Qin, and Li]{zhu2022yolo}
Daoheng Zhu, Rui Qi, Pengpeng Hu, Qianxin Su, Xue Qin, and Zhiqiang Li.
\newblock {YOLO-Rip: A modified lightweight network for Rip currents detection}.
\newblock \emph{Frontiers in Marine Science}, 9:\penalty0 930478, 2022.

\end{thebibliography}
}

\end{document}